\documentclass[10pt,twocolumn,letterpaper]{article}

\usepackage[pagenumbers]{cvpr} 

\definecolor{cvprblue}{rgb}{0.21,0.49,0.74}
\usepackage[pagebackref,breaklinks,colorlinks,allcolors=cvprblue]{hyperref}
\usepackage{multirow}        
\newcommand{\Mat}{\boldsymbol}
\newcommand{\Set}{\mathcal}

\newcommand{\real}{\mathbb{R}}

\def\tU{{\tilde{U}}}

\usepackage[accsupp]{axessibility}

\def\acronym{PUP 3D-GS}

\title{\acronym{}: Principled Uncertainty Pruning \\
for 3D Gaussian Splatting}

\author{
Alex Hanson\footnotemark[1] \hspace{2em} Allen Tu\footnotemark[1] \hspace{2em} Vasu Singla \hspace{2em} Mayuka Jayawardhana \\
Matthias Zwicker \hspace{4em} Tom Goldstein \\
[0.65em]
\textnormal{University of Maryland, College Park} \\
[0.75em]
\url{https://pup3dgs.github.io}\\
[0.75em]
}

\begin{document}
\twocolumn[{
\renewcommand\twocolumn[1][]{#1}
\maketitle
\begin{center}
    \vspace{-22pt}
    \includegraphics[width=\linewidth, trim = 0cm 0cm 0cm 19cm, clip]{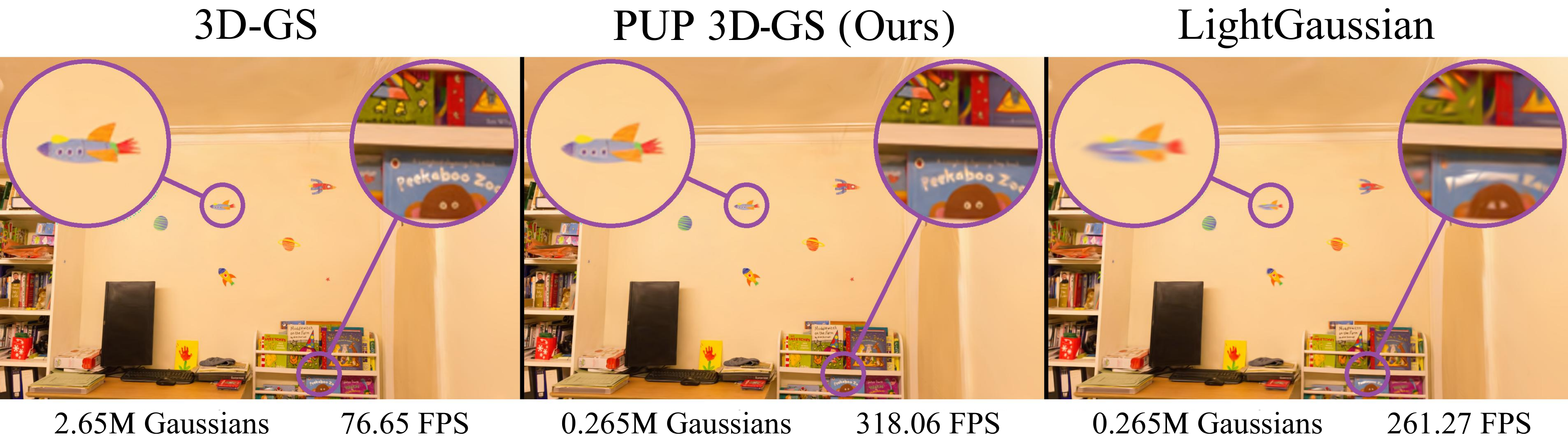}
    \vspace{-17pt}
    \captionsetup{type=figure}
    \caption{ We prune the 3D Gaussian Splatting (3D-GS) reconstruction of the Deep Blending \emph{playroom} scene from $2.65$M Gaussians to $0.265$M Gaussians using our \acronym{} pipeline, accelerating rendering speed from $76.65$ FPS to $318.06$ FPS -- a $\boldsymbol{4.15\times}$ \textbf{speed-up} on this scene-- while preserving fine details. In comparison, LightGaussian, a recent high-performing post-hoc pruning pipeline for pretrained 3D-GS models, loses substantially more fine details than \acronym{} and achieves a slower rendering speed of $261.27$ FPS.}
    \vspace{3mm}
    \label{fig:teaser}
\end{center}
}]

\makeatletter
\renewcommand\@makefnmark{\hbox{* }}
\makeatother
\footnotetext[1]{ denotes equal contribution.}

\begin{abstract}
Recent advances in novel view synthesis have enabled real-time rendering speeds with high reconstruction accuracy.
3D Gaussian Splatting (3D-GS), a foundational point-based parametric 3D scene representation, models scenes as large sets of 3D Gaussians.
However, complex scenes can consist of millions of Gaussians, resulting in high storage and memory requirements that limit the viability of 3D-GS on devices with limited resources.
Current techniques for compressing these pretrained models by pruning Gaussians rely on combining heuristics to determine which Gaussians to remove. 
At high compression ratios, these pruned scenes suffer from heavy degradation of visual fidelity and loss of foreground details. 
In this paper, we propose a principled sensitivity pruning score that preserves visual fidelity and foreground details at significantly higher compression ratios than existing approaches.
It is computed as a second-order approximation of the reconstruction error on the training views with respect to the spatial parameters of each Gaussian.
Additionally, we propose a multi-round prune-refine pipeline that can be applied to any pretrained 3D-GS model without changing its training pipeline.
After pruning $\mathit{90\%}$ of Gaussians, a substantially higher percentage than previous methods, our \acronym{} pipeline increases average rendering speed by $\mathit{3.56\times}$ while retaining more salient foreground information and achieving higher image quality metrics than existing techniques on scenes from the Mip-NeRF 360, Tanks \& Temples, and Deep Blending datasets. 
\end{abstract}
\section{Introduction}
\label{sec:intro}

Novel view synthesis aims to render views from new viewpoints given a set of 2D images. Neural Radiance Fields (NeRFs)~\cite{srinivasan2020nerf} use volume rendering to represent the 3D scene using a multilayer perceptron that can be used to render novel views. Although NeRF and its variants achieve high-quality reconstructions, they suffer from slow inference and require several seconds to render a single image. 3D Gaussian Splatting (3D-GS)~\cite{kerbl3Dgaussians} has recently emerged as a faster alternative to NeRF, achieving real-time rendering on modern GPUs and comparable image quality. It represents scenes using a large set of 3D Gaussians with independent location, shape, and appearance parameters. Since they typically consist of millions of Gaussians, 3D-GS scenes often have high storage and memory requirements that limit their viability on devices with limited resources.

Several recent works propose techniques that reduce the storage requirements of 3D-GS models, including heuristics for pruning Gaussians that have an insignificant contribution to the rendered images \cite{fan2023lightgaussian, girish2023eagles, liu2024compgs, niedermayr2023compressed, lee2023compact}. In this work, we propose a more mathematically principled approach for deciding which Gaussians to prune from a 3D-GS scene. 
We introduce a computationally feasible sensitivity score that is derived from the Hessian of the reconstructed error on the training images in a converged scene.
Following the methodology in LightGaussian~\cite{fan2023lightgaussian}, we prune the scene using our sensitivity score and then fine-tune the remaining Gaussians. Our experiments demonstrate that this process can remove $80\%$ of the Gaussians in the base 3D-GS scene while preserving image quality metrics.

Moreover, our approach enables a second consecutive round of pruning and fine-tuning. As shown in Figure~\ref{fig:teaser}, our multi-round pruning approach can remove $90\%$ of the Gaussians in the base 3D-GS scene while surpassing previous heuristics-based methods like LightGaussian in both rendering speed and image quality. Our work is orthogonal to several other methods that modify the training framework of 3D-GS or apply quantization-based techniques to reduce its disk storage. We show that our \acronym{} pipeline can be used in conjunction with these techniques to further improve performance and compress the model. 

In summary, we propose the following contributions:
\begin{enumerate}
    \item A post-hoc pruning technique, \acronym{}, that can be applied to any pretrained 3D-GS model without changing its training pipeline.
    \item A novel spatial sensitivity score that is more effective at pruning Gaussians than heuristics-based approaches.
    \item A multi-round prune-refine approach that produces higher fidelity reconstructions than an equivalent single round of prune-refining.
\end{enumerate}

\section{Related work}
\label{sec:relate}

Neural Radiance Field (NeRF) based \cite{srinivasan2020nerf} approaches use neural networks to represent scenes and perform novel-view synthesis. They produce remarkable visual fidelity but suffer from slow training and inference. Recently, 3D Gaussian Splatting (3D-GS) \cite{kerbl3Dgaussians} has emerged as an effective alternative for novel view synthesis, achieving faster inference speed and comparable visual fidelity to state-of-the-art NeRF-based approaches \cite{barron2022mipnerf360}. 

\subsection{Uncertainty Estimation}

Several works estimate uncertainty in NeRF-based approaches that arise from sources like transient objects, camera model differences, and lightning changes \cite{jin2023neu,pan2022activenerf,sabour2023robustnerf, Lyu_2024}. Other works focus on uncertainty caused by occlusion or sparsity of training views \cite{martin2021nerf}. These approaches rely on an ensemble of models \cite{sunderhauf2023density} or variational inference and KL Divergence \cite{shen2021stochastic, shen2022conditional}, requiring intricate changes to the training pipeline to model uncertainty. 

BayesRays \cite{goliBayesRays} uses Fisher information for post-hoc uncertainty quantification in NeRF-based approaches. Since NeRFs \cite{martin2021nerf} represent the scene as a continuous 3D function, they compute the Hessian over a hypothetical perturbation field to estimate uncertainty. In contrast, our approach focuses on 3D Gaussian Splats \cite{kerbl3Dgaussians} instead, and we directly compute the Fisher information using gradient information without relying on a hypothetical perturbation field. 

FisherRF \cite{jiang2023fisherrf}  also computes Fisher information for 3D Gaussian Splats; however, it only approximates the \emph{diagonal} of the Fisher matrix and uses the color parameters (DC color and spherical harmonic coefficients) of the Gaussians for post-hoc uncertainty estimation. Our approach uses the spatial mean and scaling parameters to compute a more accurate \emph{block-wise} approximation of the Fisher instead (see Section \ref{sec:method:fisher_approx}). Lastly, our work uses uncertainty estimates to prune Gaussians from the model, whereas FisherRF applies their method to perform active-view selection.

\subsection{Pruning Gaussian Splat Models}

While 3D-GS \cite{kerbl3Dgaussians} demonstrates remarkable performance, it also entails substantial storage requirements. Several recent works use codebooks to quantize and reduce storage for various Gaussian parameters \cite{lee2023compact, navaneet2023compact3d, niedermayr2023compressed}. Others use the spatial relationships between neighboring Gaussians to reduce the number of parameters \cite{lu2023scaffold, morgenstern2023compact, chen2024hac, liu2024compgs}. Although these methods tout high compression rates, they modify the underlying primitives and training framework of 3D-GS. They also do not necessarily reduce the number of primitives and are, therefore, orthogonal to our work. We apply one such technique, Vectree Quantization \cite{fan2023lightgaussian}, to further compress our pruned scenes in Section \ref{sec:vectree}.

A recent pruning-based method, Compact-3DGS \cite{lee2023compact}, proposes a learnable masking strategy to prune small, transparent Gaussians during training. EAGLES \cite{girish2023eagles}, which we apply our pipeline to in Appendix~\ref{sec:appendix:eagles}, prunes Gaussians based on the least total transmittance per Gaussian. They also begin with low-resolution images, progressively increasing image resolution to reduce Gaussian densification during training, then quantize several attributes to reduce disk storage. LightGaussian \cite{fan2023lightgaussian} computes a global significance score for each Gaussian with heuristics, uses that score to prune the least significant Gaussians, and finally uses quantization to further reduce storage requirements. 
Mini-Splatting~\cite{fang2024minisplattingrepresentingscenesconstrained} and RadSplat~\cite{niemeyer2024radsplatradiancefieldinformedgaussian} are concurrent with our work and introduce heuristics-based pruning components. However, they primarily focus on orthogonal components, such as alternative densification strategies and pretrained NeRF priors, for 3D-GS instead of pruning.
\section{Background: 3D Gaussian Splatting}
\label{sec:background}

\label{sec:method}
\begin{figure*}[tbp]
\centering
\includegraphics[width=.469\textwidth,trim={0cm 0 0 1.75cm}]{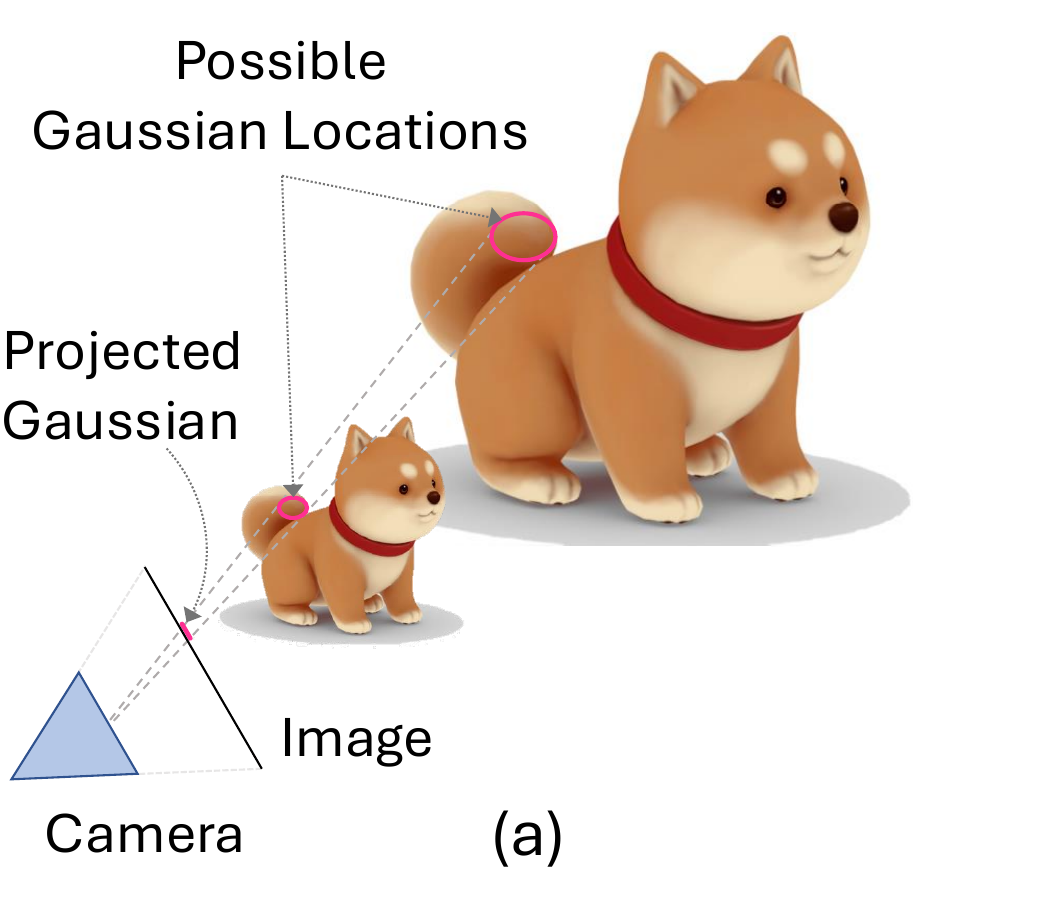}\quad\quad\quad
\includegraphics[width=.469\textwidth,trim={0cm 0 0 1.75cm}]{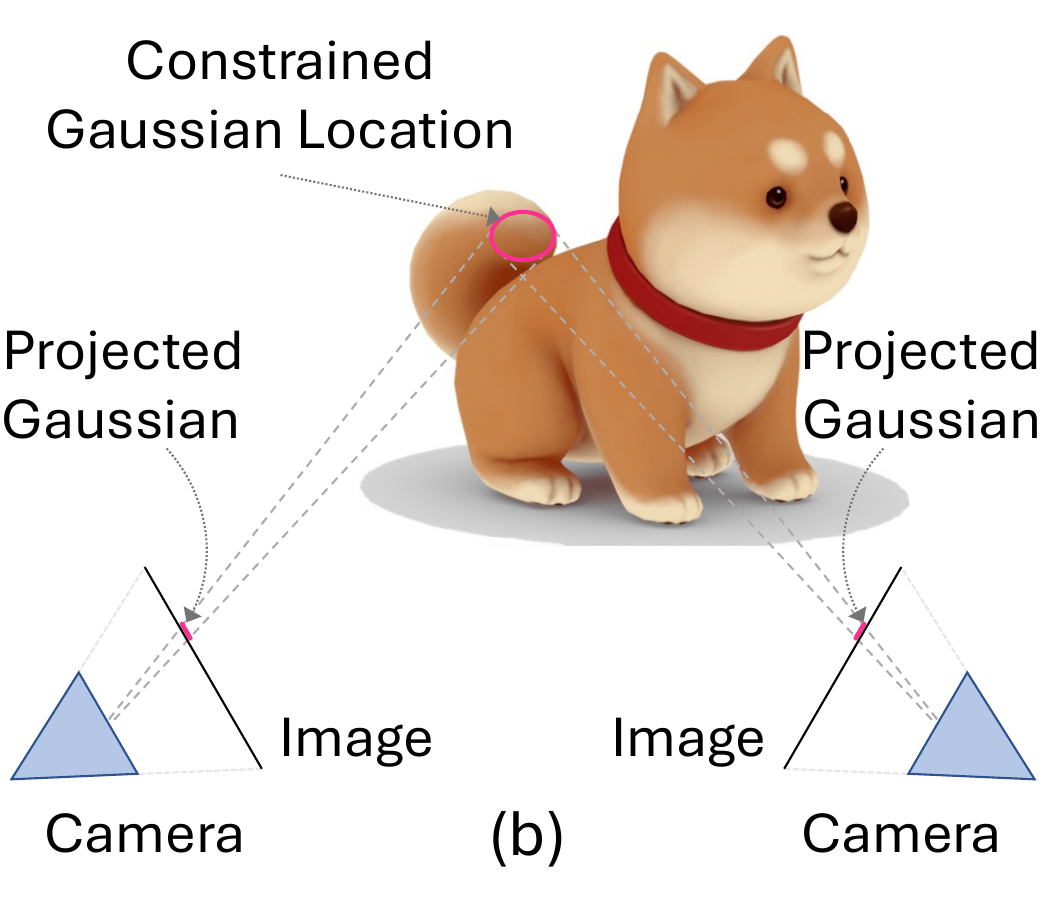} 
\vspace{-5mm} 
\caption{Spatial uncertainty arises from limited views because there are multiple possible 3D Gaussian locations that map to the same projected Gaussian in pixel space (a). This is reduced when multiple cameras observe previously unconstrained Gaussians (b).}
\vspace{-2mm}
\label{fig:scale-ambiguity}
\end{figure*}

3D Gaussian Splatting (3D-GS) is a point-based novel view synthesis technique that uses 3D Gaussians to model the scene. 
The attributes of the Gaussians are optimized over input training views given by a set of camera poses $\Set{P}_{gt} = \{\phi_i \in \real^{3 \times 4}\}_{i=1}^K$ and corresponding ground truth training images $\Set{I}_{gt} = \{\Mat{I}_i \in \real^{H \times W}\}_{i=1}^K$.
A sparse point cloud of the scene generated by Structure from Motion (SfM) over the training views is used to initialize the Gaussians.
At fixed intervals during training, a Gaussian densification step is applied to increase the number of Gaussians in areas of the model where small-scale geometry is insufficiently reconstructed. 

Each 3D Gaussian $\Set{G}_i$ is independently parameterized by position $x_i \in \real^3$, scaling $s_i \in \real^3$, rotation $r_i \in \real^4$, base color $c_i \in \real^3$, view-dependent spherical harmonics $h_i \in \real^{15\times 3}$, and opacity $\alpha_i\in\real$.
From these, we define the set of all Gaussian parameters as:
\begin{equation}
\Set{G} = \{\Set{G}_i = \{x_i, s_i, r_i, c_i, h_i, \alpha_i\}\}_{i=1}^N,
\end{equation}
where $N$ is the total number of Gaussians in the model.

During view synthesis, the scaling parameters $s_i$ and rotation parameters $r_i$ are converted into the scaling and rotation matrices $\Mat{S}_i$ and $\Mat{R}_i$. The Gaussian $\Set{G}_i$ is spatially characterized in the 3D scene by its center point, or mean position, $x_i$ and a decomposable covariance matrix $\Mat{\Sigma}_i$:
\begin{equation}
    \Set{G}_i(x_i) = e^{-\frac{1}{2}x_i^T\Mat{\Sigma}^{-1}_i x_i}, \Mat{\Sigma}_i = \Mat{R}_i\Mat{S}_i\Mat{S}_i^T\Mat{R}_i^T.
    \label{gaussian_representation}
\end{equation}
For a given camera pose $\phi$, a differentiable rasterizer renders 2D image $I_{\Set{G}}(\phi)$ by projecting all Gaussians observed from $\phi$ onto the image plane. The color of each pixel $p$ in $I_{\Set{G}}(\phi)$ is given by the blending of the $\Set{N}$ ordered Gaussians that overlap it:
\begin{equation}
    C(p)=\sum_{i\in \Set{N}} \tilde{c}_i \tilde{\alpha}_i(p) \prod_{j=1}^{i-1}(1 - \tilde{\alpha}_j(p)),
    \label{contribution}
\end{equation}
where $\tilde{c}_i$ represents the view-dependent color calculated from the camera pose $\phi$ and the optimizable per-Gaussian color $c_i$ and spherical harmonics $h_i$, and $\tilde{\alpha_i}(p)$ represents the projected Gaussian value at $p$ times the Gaussian's opacity $\alpha_i$.

The 3D-GS model is trained by optimizing the loss function $L$ via stochastic gradient descent:
\begin{equation}
    L(\Set{G} | \phi, I_{gt} ) = ||I_{\Set{G}}(\phi)-I_{gt}||_1 + L_{SSIM}(I_\Set{G}(\phi),I_{gt}),
    \label{eq:gs_loss}
\end{equation}
where the first term is a L1 residual loss and the second term is a SSIM loss.

\section{Method}

3D scene reconstruction is an inherently underconstrained problem. 
Capturing a scene as a set of posed images $(\Set{P}_{gt}, \Set{I}_{gt})$ involves projecting it onto the 2D image plane of each view.
As illustrated by Figure~\ref{fig:scale-ambiguity}, this introduces uncertainty in the locations and sizes of the Gaussians reconstructing the scene: a large Gaussian far from the camera can be equivalently modeled in pixel space by a small Gaussian close to the camera. 
In other words, Gaussians that are not perceived by a sufficient number of cameras may be able to reconstruct the input view image from a range of locations and scales.

We define \textbf{uncertainty} in 3D-GS as the amount that a Gaussian's parameters, such as its location and scale, can be perturbed without affecting the reconstruction loss over the input views.
Concretely, this is the \textbf{sensitivity} of the error over the input views to that particular Gaussian.
Given a loss function $L: \real^{\Set{G}} \to \real$ that takes the set of Gaussians $\Set{G}$ as inputs and outputs an error value over the set of training views, this sensitivity can be captured by the Hessian $\nabla^2_{\Set{G}}L$. 

In the following subsections, we demonstrate how to compute this sensitivity and use it to decide which Gaussians to prune from the model.
Directly computing the full Hessian matrix is intractable due to memory constraints. We demonstrate how to obtain a close estimate of the Hessian via a Fisher approximation in Section~\ref{sec:method:fisher_approx}.
We also find that only a block-wise approximation of the Hessian parameters is needed to quantify Gaussian sensitivity in Section~\ref{sec:method:sensitivity_score}, and that computing this over image patches is sufficient for pruning in Section~\ref{sec:method:patch_wise}.
Finally, we find that multiple rounds of pruning and fine-tuning improves our performance over one-shot pruning and fine-tuning, giving our full \acronym{} pipeline in Section~\ref{sec:method:multi_round}.

\subsection{Fisher Information Matrix}
\label{sec:method:fisher_approx}

To obtain a per-Gaussian sensitivity, we begin by taking the $L_2$ error over the input reconstruction images $I_{\Set{G}}$:
\begin{equation}
L_2 = \frac{1}{2}\sum_{\phi \in \Set{P}_{gt}}||I_{\Set{G}}(\phi) - I_{gt}||_2^2.
\end{equation}
Differentiating this twice gives us the Hessian:
\begin{equation}
\nabla_{\Set{G}}^2 L_2 = \sum_{\phi \in \Set{P}_{gt}}\nabla_{\Set{G}} I_{\Set{G}}(\phi) \nabla_{\Set{G}} I_{\Set{G}}(\phi)^T + (I_{\Set{G}}(\phi) - I_{gt})\nabla_{\Set{G}}^2 I_{\Set{G}}(\phi). \label{eq:l2_d2}
\end{equation}
On a converged 3D-GS model, the $||I_{\Set{G}} -I_{gt}||_1$ residual term of Equation~\ref{eq:gs_loss} approaches zero, causing the second order term $(I_{\Set{G}}(\phi) - I_{gt})\nabla_{\Set{G}}^2 I_{\Set{G}}$ in Equation~\ref{eq:l2_d2} to vanish:
\begin{equation}
\nabla_{\Set{G}}^2 L_2 = \sum_{\phi \in \Set{P}_{gt}}\nabla_{\Set{G}} I_{\Set{G}}(\phi) \nabla_{\Set{G}} I_{\Set{G}}(\phi)^T.
\end{equation}
This is identified as the Fisher Information matrix in related literature~\cite{goliBayesRays,jiang2023fisherrf}. We provide an explicit derivation of the Fisher Information matrix from this $L_2$ loss along with a Bayesian interpretation of our method in Appendix~\ref{sec:appendix:bayes}. Note that $\nabla_{\Set{G}} I_{\Set{G}}$ is the gradient over only the reconstructed images, so our approximation only depends on the input poses $\Set{P}_{gt}$ and not the input images $\Set{I}_{gt}$.

\subsection{Sensitivity Pruning Score}
\label{sec:method:sensitivity_score}

The full Hessian over the model parameters $\nabla_{\Set{G}}^2L \in \real^{\Set{G} \times \Set{G}}$ is quadratically large.
However, we find that using only a subset of these parameters is sufficient for an effective sensitivity pruning score. For brevity, we remove the $\Set{P}_{gt}$ terms.

To obtain sensitivity scores for each Gaussian $\Set{G}_i$, we restrict ourselves to the block diagonal of the Hessian that only captures inter-Gaussian parameter relationships.
This allows us to use each block as a per-Gaussian Hessian from which we can obtain an independent sensitivity score:
\begin{equation}
\Mat{H}_i = \nabla_{\Set{G}_i}I_{\Set{G}} \nabla_{\Set{G}_i}I_{\Set{G}}^T,
\end{equation}
where $\nabla_{\Set{G}_i}$ is the gradient with respect to the parameters of Gaussian $\Set{G}_i$. Intuitively, each block Hessian $\Mat{H}_i$ measures the isolated impact that perturbing only $\Set{G}_i$'s parameters would have on the reconstruction error.
To turn $\Mat{H}_i$ into a sensitivity score $\tU_i \in \real$, we take its log determinant:
\begin{equation}
\tU_i = \log |\Mat{H}_i| = \log |\nabla_{\Set{G}_i}I_{\Set{G}} \nabla_{\Set{G}_i} I_{\Set{G}}^T|.
\end{equation}
This captures the relative impact of all the parameters of Gaussian $\Set{G}_i$ on the reconstruction error.
Specifically, it is a relative volume measure of the basin around the Gaussian parameters $\Set{G}_i$ in the second-order approximation of the function describing their impact on the reconstruction error.

We find that we can restrict the per Gaussian parameters even further and consider only the spatial mean $x_i$ and scaling $s_i$ parameters for an effective sensitivity score. Our final sensitivity pruning score $U_i \in \real$ is:
\begin{equation}
U_i = \log |\nabla_{x_i,s_i}I_{\Set{G}} \nabla_{x_i,s_i} I_{\Set{G}}^T|.
\label{eq:uncertainty}
\end{equation}
We speculate that only the mean and scaling parameters are needed because they capture the projective geometric invariances that exist between the 3D scene and the input views as shown in Figure~\ref{fig:scale-ambiguity}. Rotations -- the remaining geometric parameters -- are excluded because they do not induce a change of 3D geometry when invariances are present.
We ablate the choice of parameters in Section~\ref{sec:ablations:shape_v_color}. 

\subsection{Patch-wise Uncertainty}
\label{sec:method:patch_wise}
The computation of the Hessian over the entire scene requires a sum over all per-pixel Fisher approximations of the reconstructed input views.
However, we empirically observe that sensitivity scores computed over image patches are highly correlated with those computed over individual pixels.
Specifically, we compute the Fisher information approximation on image patches by rendering images at lower resolution and then computing the Fisher approximation on each of their pixels. 
Then, we obtain the scene-level Hessian by summing the patch-wise Fisher approximations over all views.
We use $4\times4$ image patches in our experiments and ablate our choice of patch size in Appendix~\ref{sec:appendix:patch_size}.

\subsection{Multi-Round Pipeline}
\label{sec:method:multi_round}
Similar to LightGaussian~\cite{fan2023lightgaussian}, we prune the model and then fine-tune it without further Gaussian densification.
For brevity, we will refer to this process as \textbf{prune-refine}.
We find that, in many circumstances, the model is able to recover the small $||I_{\Set{G}} - I_{gt}||_1$ residual after fine-tuning, allowing us to repeat prune-refine over multiple rounds.
We empirically notice that multiple rounds of prune-refine outperforms an equivalent single round of prune-refine.
Details of this evaluation are in Section~\ref{sec:ablations:multi_round}.
\begin{figure*}[t]
  \includegraphics[width=\linewidth]{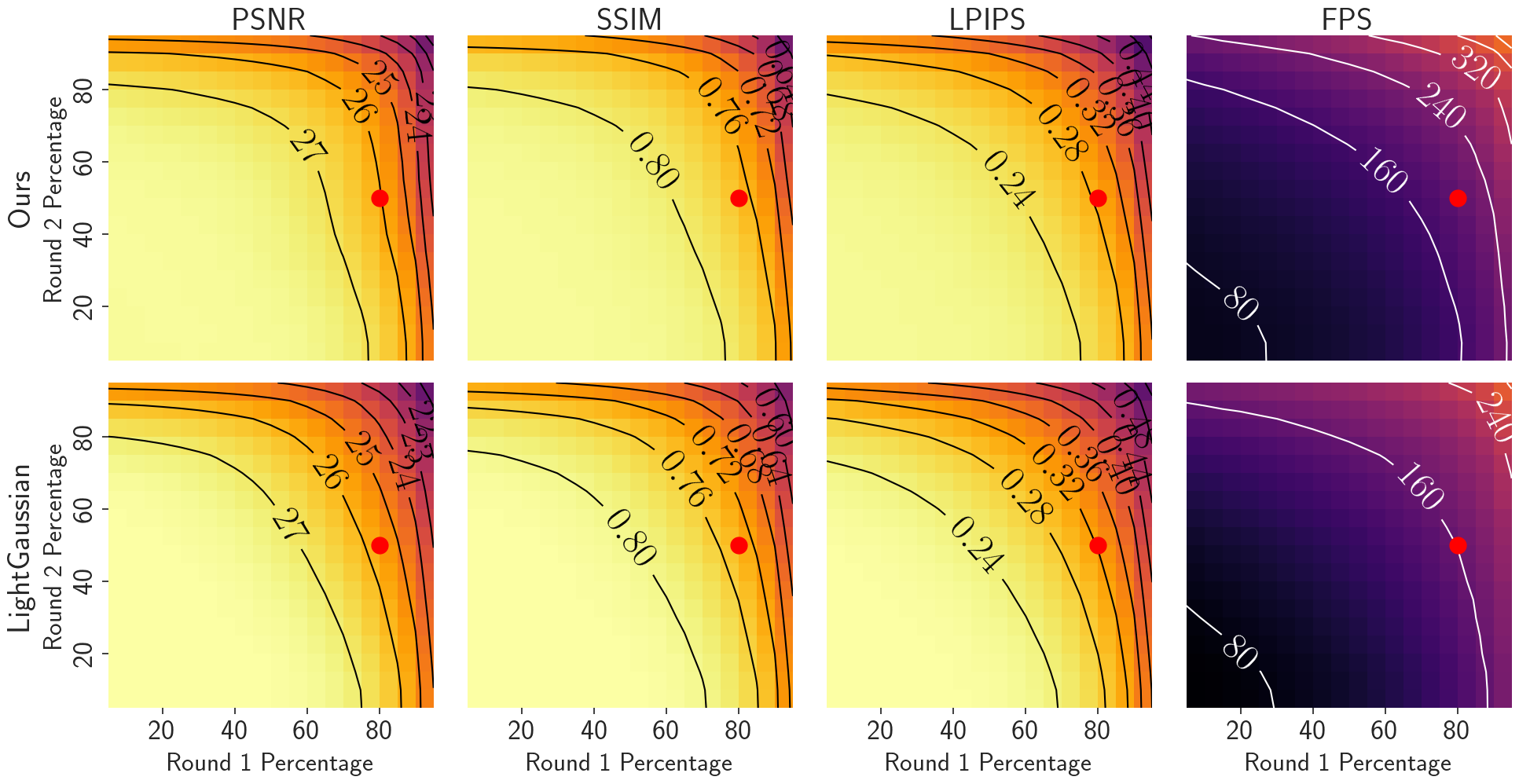}
  \vspace{-7mm}
  \caption{The average metrics over all scenes in the Mip-NeRF 360 dataset after pruning with different percentages in our two-round pipeline. PSNR and SSIM decrease and LPIPS and FPS increase with more pruning. Per-round percentages are selected in $5\%$ intervals and the model is fine-tuned for $5,000$ iterations in each round. The red dots at $(80\%,50\%)$ represent our percentages for $90\%$ compression.
  }
  \label{fig:mipnerf_heatmap}
  \vspace{-2mm}
\end{figure*}

\section{Experiments}
\label{sec:experiments}

\subsection{Datasets}
We evaluate our approach on the same challenging, real world scenes as 3D-GS \cite{kerbl3Dgaussians}. All nine scenes from the Mip-NeRF 360 dataset~\cite{barron2022mipnerf360}, which consists of five outdoor and four indoor scenes that each contain a complex central object or viewing area and a detailed background, are used. Two outdoor scenes, \emph{truck} and \emph{train}, are taken from the Tanks \& Temples dataset~\cite{Knapitsch2017tandt}, and two indoor scenes, \emph{drjohnson} and \emph{playroom}, are taken from the Deep Blending dataset~\cite{DeepBlending2018}. For consistency, we use the COLMAP~\cite{schonberger2016colmap} camera pose estimates that were provided by the creators of the datasets.

\subsection{Implementation Details}
Our method, \acronym{}, can be applied to any 3D-GS model. 
We implement the Hessian computation as a CUDA kernel in the original 3D-GS codebase \cite{kerbl3Dgaussians}, then adapt the pruning and refining framework from LightGaussian~\cite{fan2023lightgaussian} to accommodate our uncertainty estimate and multi-round pruning method.
Computing pruning scores across the entire training view set takes seconds.
For a fair comparison, we run LightGaussian and our pipeline on the same pretrained 3D-GS scenes. 
Rendering speeds are collected using a Nvidia RTXA4000 GPU in frames per second (FPS).

\subsection{Results}

\begin{figure*}[t]
  \includegraphics[width=\linewidth,trim = 0cm 0cm 0cm 4cm, clip]{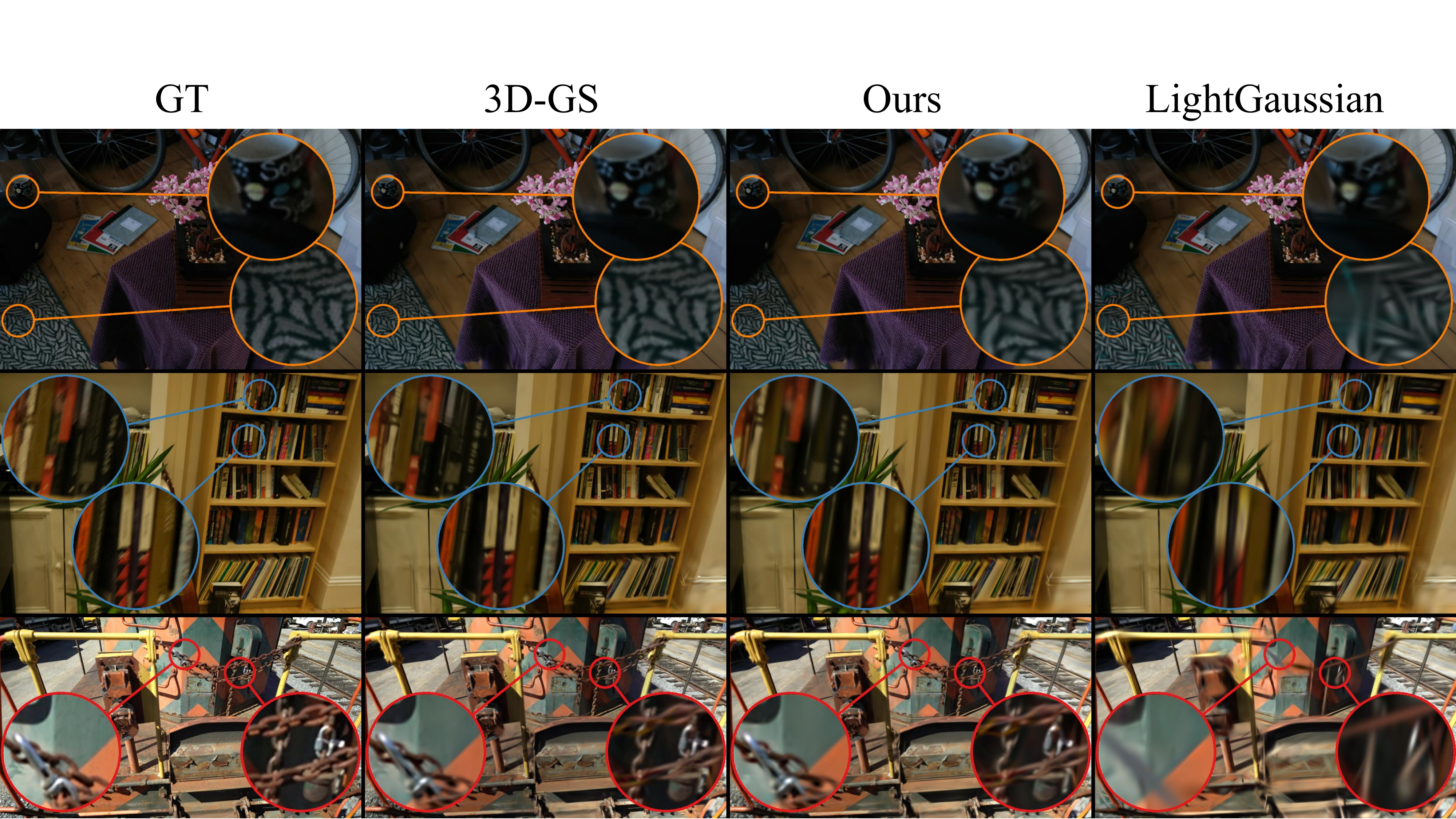}
  \vspace{-6mm}
  \caption{Visual comparison after two rounds of prune-refine using our and LightGaussian's methods. Top: \emph{bonsai} from Mip-NeRF 360. Middle: \emph{room} from Mip-NeRF 360. Bottom: \emph{train} from Tanks \& Temples. Additional visualizations are presented in Appendix~\ref{sec:appendix:scene_visuals}.}
  \vspace{-2mm}
  \label{fig:experiments_figure}
\end{figure*}

\subsubsection{Analysis of \acronym{} Pipeline}
\label{sec:pipeline}
The efficacy of our \acronym{} pipeline is demonstrated by Table~\ref{tab:pipeline_eval}, where we record the mean of each metric over the Mip-NeRF 360 dataset. Pruning $80\%$ of Gaussians from the original 3D-GS model more than doubles its FPS while degrading PSNR, SSIM, and LPIPS. Fine-tuning the pruned model for 5,000 iterations recovers the image quality metrics with only a slight decrease in rendering speed. Pruning $50\%$ of the remaining Gaussians, such that a total of $90\%$ of Gaussians are pruned from the base 3D-GS scene, causes another drop in the image quality metrics. However, we recover most of this degradation by fine-tuning this significantly smaller model for another 5,000 iterations. Rendering speed is more than tripled over the 3D-GS model.

\begin{table}[h]
\caption{Mean PSNR, SSIM, LPIPS, FPS, and point cloud size for the Mip-NeRF 360 dataset at each stage of our \acronym{} pipeline. The operation in each row is applied \textbf{cumulatively} to all of the following rows.}
\centering
\resizebox{\columnwidth}{!}{
\begin{tabular}{lccccc}
\toprule
Methods & PSNR$\uparrow$ & SSIM$\uparrow$ & LPIPS$\downarrow$ & FPS$\uparrow$ & Size (MB)$\downarrow$ \\
\midrule
3D-GS & 27.47 & 0.8123 & 0.2216 & 63.88 & 746.46 \\
+ Prune $80\%$ & 21.00 & 0.7075 & 0.3075 & 167.93 & 149.29 \\
+ Refine & 26.97 & 0.7991 & 0.2444 & 148.28 & 149.29 \\
+ Prune $50\%$ & 22.63 & 0.7021 & 0.3316 & 241.97 & 74.65 \\
+ Refine & 26.67 & 0.7862 & 0.2719 & 204.81 & 74.65 \\
\midrule
\bottomrule
\end{tabular}}
\label{tab:pipeline_eval}
\end{table}

\subsubsection{Comparison with LightGaussian}
\label{sec:experiments:compare_methods}

Figure~\ref{fig:mipnerf_heatmap} compares our spatial uncertainty estimate against LightGaussian's global significance score at different per-round pruning percentages in our two-round prune refine pipeline. After each round of pruning, the models are fine-tuned for $5,000$ iterations for a total of $10,000$ iterations. The per-dataset means of each metric are illustrated as heatmaps. As illustrated by the contour lines, our method consistently outperforms LightGaussian across all metrics and pruning percentage permutations.

We choose to prune $80\%$ then $50\%$ of the Gaussians in the first and second rounds of pruning to optimize image quality and rendering speed at $90\%$ pruning, a significantly higher compression ratio than previous methods. Table~\ref{tab:compare_methods} compares our $90\%$ pruning results against LightGaussian's using the per-dataset mean of each metric. We show that our spatial uncertainty estimate outperforms LightGaussian across nearly every metric -- the lone exception is PSNR in the Tanks \& Temples dataset, where we are outperformed by LightGaussian by $0.36$ points. Our choice of pruning percentages is further ablated in Section~\ref{sec:percentages}. 

\begin{table}[t]

\caption{Comparison of our \acronym{} pipeline against LightGaussian \cite{fan2023lightgaussian}. In both methods, we prune-refine $80\%$ of the Gaussians from the 3D-GS model and then $50\%$ of the remaining Gaussians for a total of $90\%$. The final sizes are identical. \acronym{} increases the rendering speed of 3D-GS by $3.56\times$ on average. Per-scene metrics for each dataset are recorded in Appendix~\ref{sec:appendix:scene_eval}.
}
\vspace{-1.5mm}
\centering
\resizebox{\columnwidth}{!}{
\begin{tabular}{ccccccc}
\toprule
Datasets & Methods &  PSNR$\uparrow$ & SSIM$\uparrow$ & LPIPS$\downarrow$ & FPS$\uparrow$ & Size (MB)$\downarrow$ \\
\midrule
\multirow{3}{*}{MipNeRF-360} & 3D-GS & 27.47 & 0.8123 & 0.2216 & 64.07 & 746.46 \\
 & LightGaussian & 26.28 & 0.7622 & 0.3054 & 162.12 & 74.65 \\
 & Ours & \textbf{26.67} & \textbf{0.7862} & \textbf{0.2719} & \textbf{204.81} & 74.65 \\
 \midrule
\multirow{3}{*}{Tanks \& Temples}  & 3D-GS & 23.77 & 0.8458 & 0.1777 & 97.86 & 433.24 \\
 & LightGaussian & \textbf{23.08} & 0.7950 & 0.2634 & 329.03 & 43.33 \\
 & Ours & 22.72 & \textbf{0.8013} & \textbf{0.2441} & \textbf{391.10} & 43.33 \\
 \midrule
\multirow{3}{*}{Deep Blending}  & 3D-GS & 28.98 & 0.8816 & 0.2859 & 66.79 & 699.19 \\
 & LightGaussian & 28.51 & 0.8675 & 0.3292 & 234.10 & 69.92 \\
 & Ours & \textbf{28.85} & \textbf{0.8810} & \textbf{0.3015} & \textbf{301.43} & 69.92 \\
\midrule
 \bottomrule
\end{tabular}}
\label{tab:compare_methods}
\vspace{-2.5mm}
\end{table}

In Figures~\ref{fig:teaser}~and~\ref{fig:experiments_figure}, we report qualitative results on scenes taken from each dataset. The magnified regions demonstrate that our method retains many fine foreground details that are not preserved by LightGaussian. Error residual visualizations with respect to the ground truth images and original 3D-GS renders are available in Appendix~\ref{sec:appendix:l1_gt_error}; visualizations of other scenes are provided in Appendix~\ref{sec:appendix:scene_visuals}.

\begin{figure*}[t]
  \includegraphics[width=\linewidth]{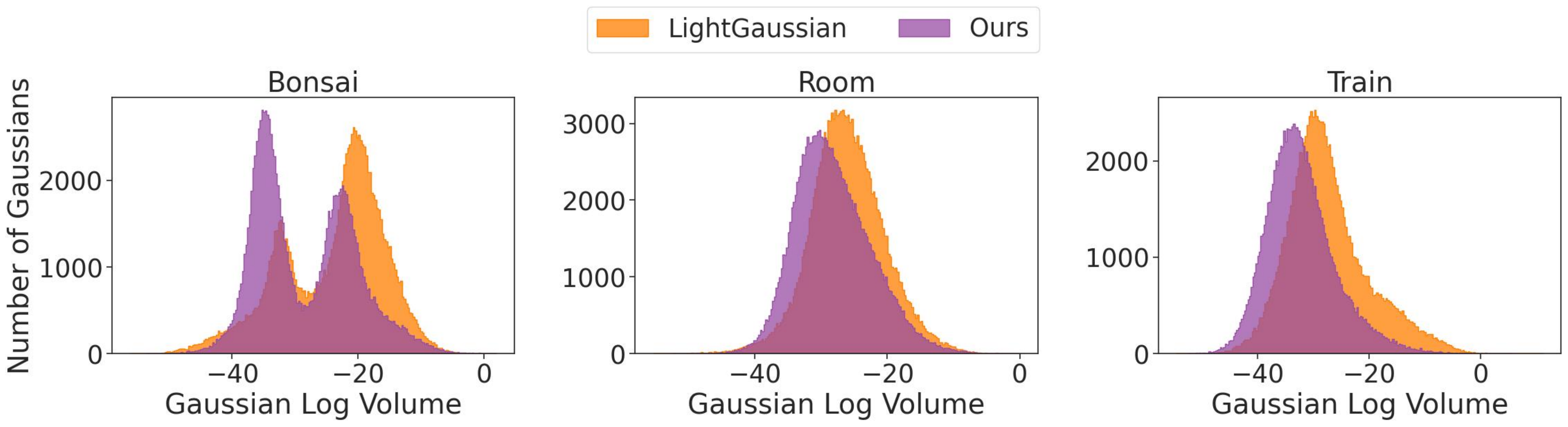}
  \vspace{-6mm}
  \caption{Histograms of the distribution of Gaussians over the log of their volumes for the \emph{bonsai}, \emph{room}, and \emph{train} scenes after two rounds of prune-refine. \acronym{} retains smaller Gaussians than LightGaussian, consistent with its higher rendering speed and visual fidelity. }
  \label{fig:experiments_histogram}
  \vspace{-2mm}
\end{figure*}

We speculate that the visibility score in LightGaussian's importance sampling heuristic is biased towards retaining larger Gaussian because they have a higher probability of being visible in more pixels across the training views. In Figure~\ref{fig:experiments_histogram}, we plot the distributions of the log determinants of the Gaussian covariances over the scenes visualized in Figure~\ref{fig:experiments_figure}. The distributions skew more heavily towards larger Gaussians in the scenes pruned with LightGaussian's pipeline than ours. We address other pruning methods \cite{lee2023compact, niemeyer2024radsplatradiancefieldinformedgaussian, fang2024minisplattingrepresentingscenesconstrained} in Appendix~\ref{sec:appendix:other_pruning}.

\subsubsection{Vectree Quantization}
\label{sec:vectree}

We apply Vectree Quantization -- a compression method that is orthogonal to pruning -- from the LightGaussian framework~\cite{fan2023lightgaussian} to further compress our pruned models. Our final models are smaller than the unpruned 3D-GS scenes by an average of $51.70\times$, $52.56\times$, and $51.06\times$ for the Mip-NeRF 360, Tanks \& Temples, and Deep Blending datasets. Detailed results are in Table~\ref{tab:quantization}. We also apply our method to EAGLES~\cite{girish2023eagles}, another orthogonal method, in Appendix~\ref{sec:appendix:eagles}.

\begin{table}[h!]
\vspace{1mm}
\caption{Vectree Quantization comparison with LightGaussian after pruning to $90\%$. \acronym{} outperforms LightGaussian in terms of image quality, rendering speed, and now size.
}
\centering

\resizebox{\columnwidth}{!}{
\begin{tabular}{ccccccc}
\toprule
\textbf{Datasets} & \textbf{Methods} & PSNR$\uparrow$ & SSIM$\uparrow$ & LPIPS$\downarrow$ & FPS$\uparrow$ & Size (MB)$\downarrow$ \\
\midrule
\multirow{3}{*}{Mip-NeRF 360}  & 3D-GS & 27.47 & 0.8123 & 0.2216 & 64.07 & 746.46 \\
 & LightGaussian & 24.65 & 0.7302 & 0.3341 & 162.34 & 14.44 \\
 & Ours & \textbf{24.93} & \textbf{0.7584} & \textbf{0.2988} & \textbf{205.97} & \textbf{14.44} \\
\midrule
\multirow{3}{*}{Tanks \& Temples}  & 3D-GS & 23.77 & 0.8458 & 0.1777 & 97.86 & 433.24 \\
 & LightGaussian & \textbf{21.88} & 0.7679 & 0.2886 & 331.47 & 8.53 \\
 & Ours & 21.61 & \textbf{0.7787} & \textbf{0.2670} & \textbf{389.18} & \textbf{8.49} \\
\midrule
\multirow{3}{*}{Deep Blending}  & 3D-GS & 28.98 & 0.8816 & 0.2859 & 66.79 & 699.19 \\
 & LightGaussian & 27.90 & 0.8586 & 0.3403 & 233.28 & 13.38 \\
 & Ours & \textbf{28.24} & \textbf{0.8735} & \textbf{0.3108} & \textbf{300.15} & \textbf{13.30} \\
 \midrule
 \bottomrule
\end{tabular}}
\vspace{-7mm}
\label{tab:quantization}
\end{table}

\section{Ablations}
\label{sec:ablations}

\subsection{Multi-Round Pruning}
\label{sec:ablations:multi_round}

A core component of our pipeline is multiple rounds of prune-refine. In Table~\ref{tab:ablation_multi_step}, we compare the mean image quality metrics and FPS over the Mip-NeRF 360 scenes when we prune using our two step prune-refine method against pruning $90\%$ of Gaussians in a single round with an equivalent number of fine-tuning steps. Two-round pruning produces better measurements across all metrics. 

\begin{figure}[t]
  \includegraphics[width=\columnwidth]{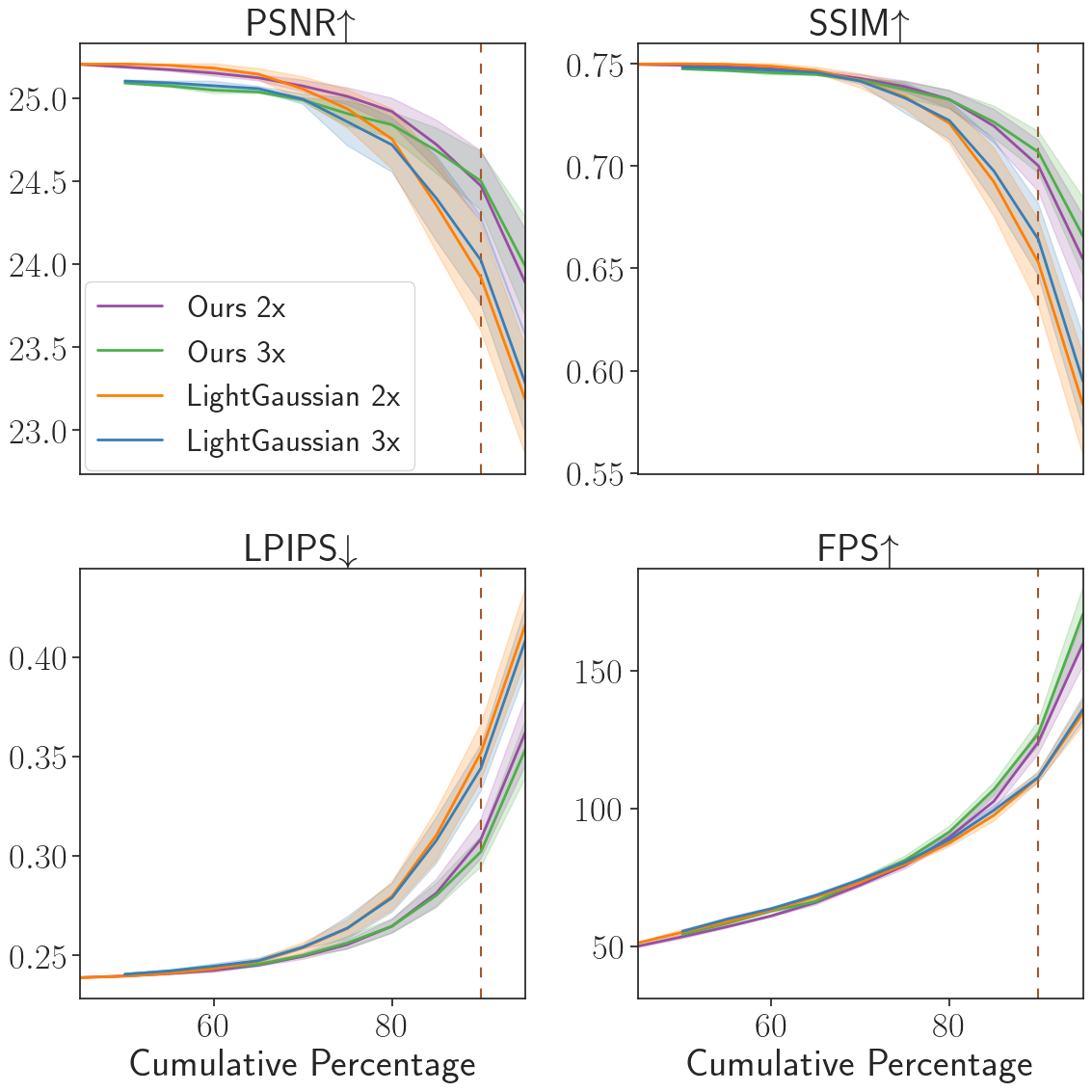}
  \vspace{-4.725mm}
  \caption{Summary results on the MipNeRF-360 \textit{bicycle} scene for $2\times$ and $3\times$ prune-refine rounds. We plot the mean and standard deviation of $2\times$ for all cumulative pruning percents in Figure~\ref{fig:mipnerf_heatmap}, computed via cubic interpolation, then do the same for $3\times$ with $10\%$ pruning intervals. The dotted red line denotes our target $90\%$ cumulative percentage; per-round results are ablated in Figure~\ref{fig:90_plot}.
  }
  \vspace{-4.2mm}
  \label{fig:multiple_rounds}
\end{figure}

The number of pruning percentage permutations increases exponentially as the number of rounds increases, so we use interpolation to compare two against three rounds of pruning on the Mip-NeRF 360 \textit{bicycle} scene in Figure~\ref{fig:multiple_rounds}. Three-round pruning is substantially worse at lower percentages and only slightly better at higher percentages. We choose the simpler two-round approach for our pipeline because both configurations outperform LightGaussian by a much larger margin at $90\%$.

\begin{table}[t]
\caption{One step of pruning $90\%$ of Gaussians from 3D-GS and then fine-tuning for 10,000 iterations, versus pruning $80\%$ then $50\%$ of Gaussians and fine-tuning for 5,000 iterations in each step. In both methods, $90\%$ of Gaussians are pruned cumulatively and the model is fine-tuned for 10,000 total steps. Our multi-round approach outperforms the one-round approach across all metrics.}
\centering
\resizebox{\columnwidth}{!}{
\begin{tabular}{c*{5}{c}}
    \toprule
    \multirow{2}{*}{\textbf{Methods}} & \multicolumn{5}{c}{Mip-NeRF 360} \\
    \cmidrule{2-6} & PSNR$\uparrow$ & SSIM$\uparrow$ & LPIPS$\downarrow$ & FPS$\uparrow$ & Size (MB)$\downarrow$ \\
    \midrule
    3D-GS            & 27.47 & 0.8123 & 0.2216 & 95.59 & 746.46 \\
     Prune $90\% + 10$K Fine-tune & 26.12 & 0.7761 & 0.2807 & 189.95 & 74.65 \\
     Prune $80\% + 50\%$ ($5$K Each) & \textbf{26.67} & \textbf{0.7862} & \textbf{0.2719} & \textbf{204.81} & 74.65   \\
    \midrule
    \bottomrule
\end{tabular}}
\label{tab:ablation_multi_step}
\vspace{-2mm}
\end{table}

\subsection{Per-Round Pruning Percentages}
\label{sec:percentages}
\begin{figure}[t]
   \includegraphics[width=\columnwidth]{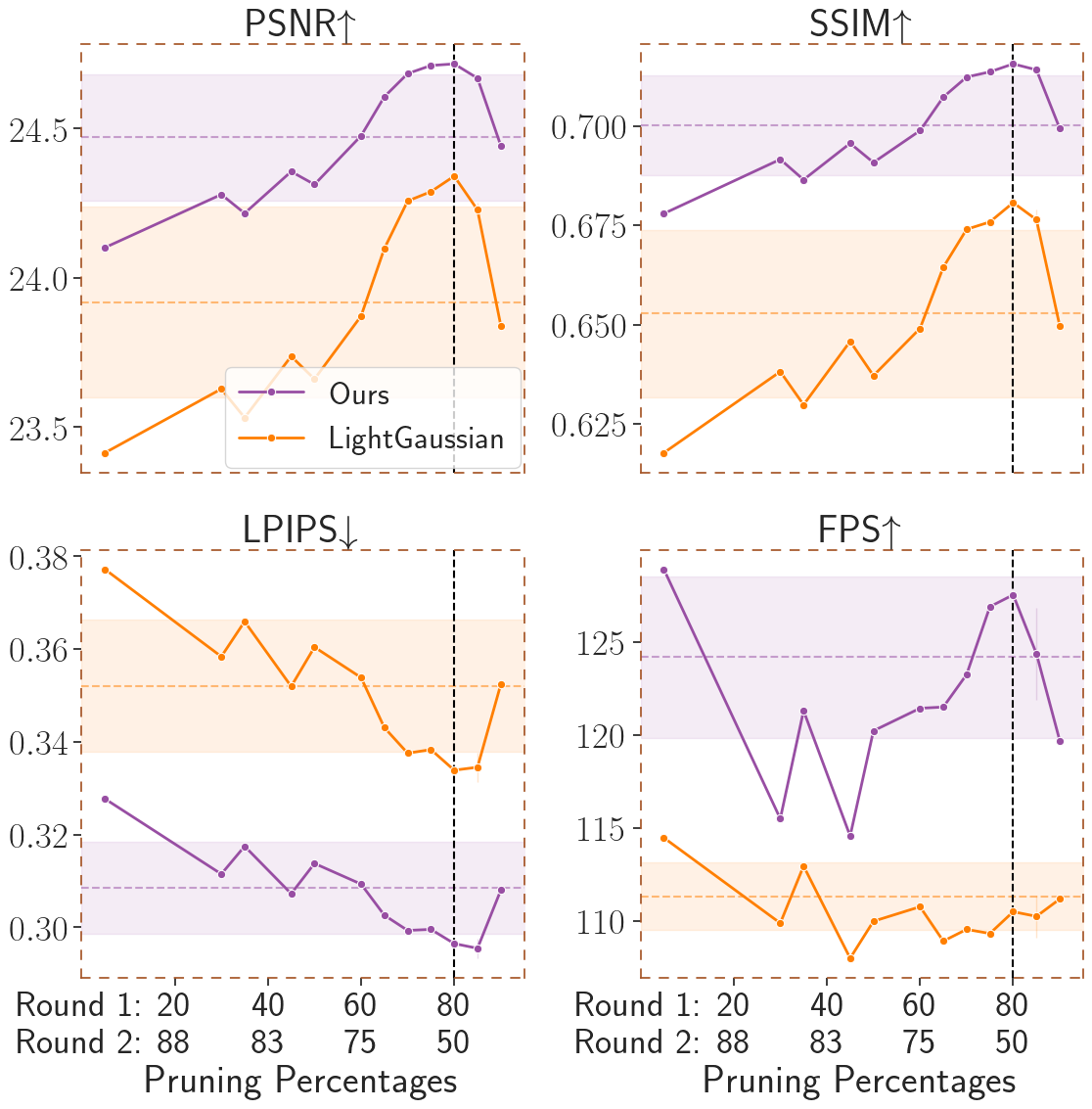}
  \vspace{-6mm}
  \caption{Results for the MipNeRF-360 \textit{bicycle} scene when pruning approximately $90\%$ of Gaussians with varying per-round percentages in our two-round pipeline. The dotted black line denotes our chosen pruning percentages of $80\%$ then $50\%$; the $90\%$ means from Figure~\ref{fig:multiple_rounds} are also shown as dotted lines. Appendix~\ref{sec:appendix:90_ablation} reports the average metrics across all scenes in the MipNeRF-360 dataset.
  }
  \vspace{-5.37mm}
  \label{fig:90_plot}
\end{figure}

We choose to prune $90\%$ of Gaussians, substantially more than previous methods, because preserving visual fidelity at extreme compression ratios is challenging. Figure~\ref{fig:90_plot} plots the metrics for each permutation of per-round pruning percentages that results in approximately $90\%$ total pruning. Pruning $80\%$ then $50\%$ of the Gaussians in the first and second rounds optimizes image quality and rendering speed at exactly $90\%$ total pruning. Our method outperforms LightGaussian across all metrics and percentage permutations.

\subsection{Spatial vs. Color Parameters}
\label{sec:ablations:shape_v_color}

Another component of our method is restricting our sensitivity pruning score to only the mean and scale parameters of each Gaussian. We choose to exclude rotations, the remaining geometric parameters, because they do not induce a change in 3D geometry when invariances are present. Furthermore, including rotation parameters produces $10\times10$ Fisher Information matrices that are $2.78\times$ larger than the $6\times6$ mean and scale parameter-only matrices used in our method, inducing significant additional computational overhead. It is no longer possible to run the pruning pipeline on the Nvidia RTXA4000 GPU used for our other experiments due to memory constraints, necessitating  more expensive hardware like the Nvidia RTXA5000.

Table~\ref{tab:ablation_rgb} shows that image quality decreases if we include the rotation parameters in our mean and scale parameter sensitivity score. Image quality also decreases if we use the RGB color parameters of the Gaussians to compute our sensitivity score instead. We speculate that the lower performance of the RGB sensitivity score is due to its similarity with LightGaussian's visibility score -- the change in the L2 error over the training images with respect to perturbations in the RGB value of a Gaussian correlates with its visibility across the training views.

\begin{table}[h!]
\vspace{0mm}
\caption{Results from our two step prune-refine pipeline when computing sensitivity via our score, our score with rotation parameters, and RGB color parameters. Our spatial score outperforms the spatial with rotation and RGB color scores.}
\centering
\resizebox{\columnwidth}{!}{
\begin{tabular}{c*{5}{c}}
    \toprule
    \multirow{2}{*}{\textbf{Methods}} & \multicolumn{5}{c}{Mip-NeRF 360} \\
    \cmidrule{2-6} & PSNR$\uparrow$ & SSIM$\uparrow$ & LPIPS$\downarrow$ & FPS$\uparrow$ & Size (MB)$\downarrow$ \\
    \midrule
    3D-GS            & 27.47 & 0.8123 & 0.2216 & 95.59 & 746.46 \\
    Ours        & 26.67 & \textbf{0.7862} & \textbf{0.2719} & \textbf{204.81}  & 74.65 \\
    Ours + Rotation      & \textbf{26.79} & 0.7861 & 0.2726 & 177.30 & 74.65 \\
    RGB Sensitivity  & 26.40 & 0.7798 & 0.2769 & 195.76 & 74.65 \\
    \midrule
    \bottomrule
\end{tabular}}
\vspace{-4mm}
\label{tab:ablation_rgb}
\end{table}

\section{Limitations}
\label{sec:limitations}

A limitation of our approach is that it assumes that the scene is converged because the Fisher approximation requires a small L1 residual to be accurate.
We find that a small residual is preserved even after removing $80\%$ of the Gaussians with prune-refine, allowing for a second round of pruning, as shown in Section \ref{sec:method:multi_round}. Another limitation is that the memory requirement for computing the Fisher matrices is proportional to $N\times36$, where $N$ is the total number of Gaussians. We do not find this size to be prohibitive for our post-hoc pruning pipeline on 16GB Nvidia RTXA4000 or larger GPUs. Appendix~\ref{sec:appendix:next_steps} discusses potential directions for future research related to \acronym{}.

\section{Conclusion}

In this work, we propose \acronym{}: a new post-hoc pipeline for pruning pretrained 3D-GS models without changing their training pipelines. 
It uses a novel, mathematically principled approach for choosing which Gaussians to prune by assessing their sensitivity to the reconstruction error over the training views.
This sensitivity pruning score is derived from a computationally tractable Fisher approximation of the Hessian over that reconstruction error.
We use our sensitivity score to prune Gaussians from the reconstructed scene, then fine-tune the remaining Gaussians in the model over multiple rounds. 
After pruning $90\%$ of Gaussians, \acronym{} increases average rendering speed by $3.56\times$, while retaining more salient foreground information and achieving higher image quality metrics than existing techniques on scenes from the Mip-NeRF 360, Tanks \& Temples, and Deep Blending datasets.

\section{Acknowledgements}
This research is based upon work supported by the Office of the Director of National Intelligence (ODNI), Intelligence Advanced Research Projects Activity (IARPA), via IARPA R\&D Contract No. 140D0423C0076. The views and conclusions contained herein are those of the authors and should not be interpreted as necessarily representing the official policies or endorsements, either expressed or implied, of the ODNI, IARPA, or the U.S. Government. The U.S. Government is authorized to reproduce and distribute reprints for Governmental purposes notwithstanding any copyright annotation thereon. Additional support was provided by ONR MURI program and the AFOSR MURI program. Commercial support was provided by Capital One Bank, the Amazon Research Award program, and Open Philanthropy. Zwicker was additionally supported by the National Science Foundation (IIS-2126407). Goldstein was additionally supported by the National Science Foundation (IIS-2212182) and by the NSF TRAILS Institute (2229885).

{
    \small
    \bibliographystyle{ieeenat_fullname}
    \bibliography{main}
}

\clearpage
\appendix

\section{Appendix}

\subsection{Fisher Information Derivation: A Bayesian Interpretation of \acronym{}}
\label{sec:appendix:bayes}
The Fisher Information is the variance of the score:
\begin{equation}
I(\theta) = E_{\theta}[(\nabla_{\theta} \log p(x|\theta))^2].
\end{equation}
Lemma 5.3 from~\cite{lehmann2006theory} gives that this is equivalent to:
\begin{equation}
I(\theta) = E_{\theta}[ -\nabla_{\theta} \nabla_{\theta} \log p(x|\theta)],
\end{equation}
assuming $\log p(x|\theta)$ is twice differentiable and with certain regularity conditions.

To start, we reformulate our $L_2$ objective as a log-likelihood:
\begin{equation}
-\log p(\mathcal{I}|\Phi,\mathcal{G})= E_{(I,\phi)\sim(\mathcal{I},\Phi)}[(I - I_{\mathcal{G}}(\phi))^T (I - I_{\mathcal{G}}(\phi))],
\end{equation}

where $\mathcal{I}$ is the set of ground truth image, $\Phi$ is the set of their corresponding poses, $\mathcal{G}$ are the model Gaussian parameters, $I_{\mathcal{G}}(\phi)$ is the rendering function for pose $\phi$, and we assume that the error follows a Gaussian distribution.

We can take the Laplace approximation of the log of the posterior distribution over the model parameters $\mathcal{G}$ on the converged scene parameters $\hat{\mathcal{G}}$ as:
\vspace{-1mm}
\begin{equation}
-\log p(\mathcal{G}|\mathcal{I},\Phi) \approx -\log p(\hat{\mathcal{G}}|\mathcal{I},\Phi) + \frac{1}{2}(\mathcal{G} - \hat{\mathcal{G}})H(\hat{\mathcal{G}})(\mathcal{G} - \hat{\mathcal{G}}),
\end{equation}
\vspace{-1mm}
where:
\begin{equation}
H(\hat{\mathcal{G}}) = -\nabla_{\mathcal{G}}\nabla_{\mathcal{G}} \log p(\hat{\mathcal{G}}|\mathcal{I},\Phi).
\end{equation}

If we assume a uniform prior, then our claimed Fisher Information matrix from Section ~\ref{sec:method:fisher_approx} is precisely the Hessian $H(\hat{\mathcal{G}})$ of this posterior.

From this formulation of our Fisher Information matrix, Proposition 3.5 from~\cite{kirsch2022unifying} gives that the log determinant of the Fisher as the entropy of the second order approximation of $p(\mathcal{G}|\mathcal{I},\Phi$) around $\hat{\mathcal{G}}$. If we restrict the posterior to a particular Gaussian's parameters $\mathcal{G}_i$, giving $p(\mathcal{G}_i|\mathcal{I},\Phi) $, the log determinant of the block diagonal element corresponding to this Gaussian is a measure of entropy for that particular Gaussian. This interpretation gives our pruning scores as a ranking of the Gaussians by their entropy on this posterior.

\subsection{Ablation on Patch Size}
\label{sec:appendix:patch_size}

We ablate our choice of patch size in Table~\ref{tab:patch_size}. Notice that, although the $4\times4$ patches that we use in our experiments produce slightly better image quality metrics, the $2\times2$ and $8\times8$ patches also produce similar results.

\begin{table}[h!]
\caption{Mean PSNR, SSIM, LPIPS, FPS, and point cloud size for the Mip-NeRF 360 dataset using our sensitivity score computed with $2\times2$, $4\times4$ and $8\times8$ patches.}
\centering

\resizebox{\columnwidth}{!}{
\begin{tabular}{l*{5}{c}}
    \toprule
    \multirow{2}{*}{\textbf{Methods}} & \multicolumn{5}{c}{Mip-NeRF 360} \\
    \cmidrule{2-6} & PSNR$\uparrow$ & SSIM$\uparrow$ & LPIPS$\downarrow$ & FPS$\uparrow$ & Size (MB)$\downarrow$ \\
    \midrule
    3D-GS        & 27.47 & 0.8123 & 0.2216 & 83.88 & 746.46 \\
    $2\times2$ & 26.46 & \textbf{0.7869} & 0.2742 & \textbf{218.59} & 74.65\\
    $4\times4$ (Ours) & \textbf{26.67} & 0.7862 & \textbf{0.2719} & 204.81 & 74.65\\
    $8\times8$ &  26.53 & 0.7775 & 0.2780 & 189.23 & 74.65\\
    \midrule
    \bottomrule
\end{tabular}}
\label{tab:patch_size}

\end{table}

\subsection{Ablation on Per-Round Pruning Percentages}
\label{sec:appendix:90_ablation}
\begin{figure}[t]
    \vspace{-1.2mm}
   \includegraphics[width=\columnwidth]{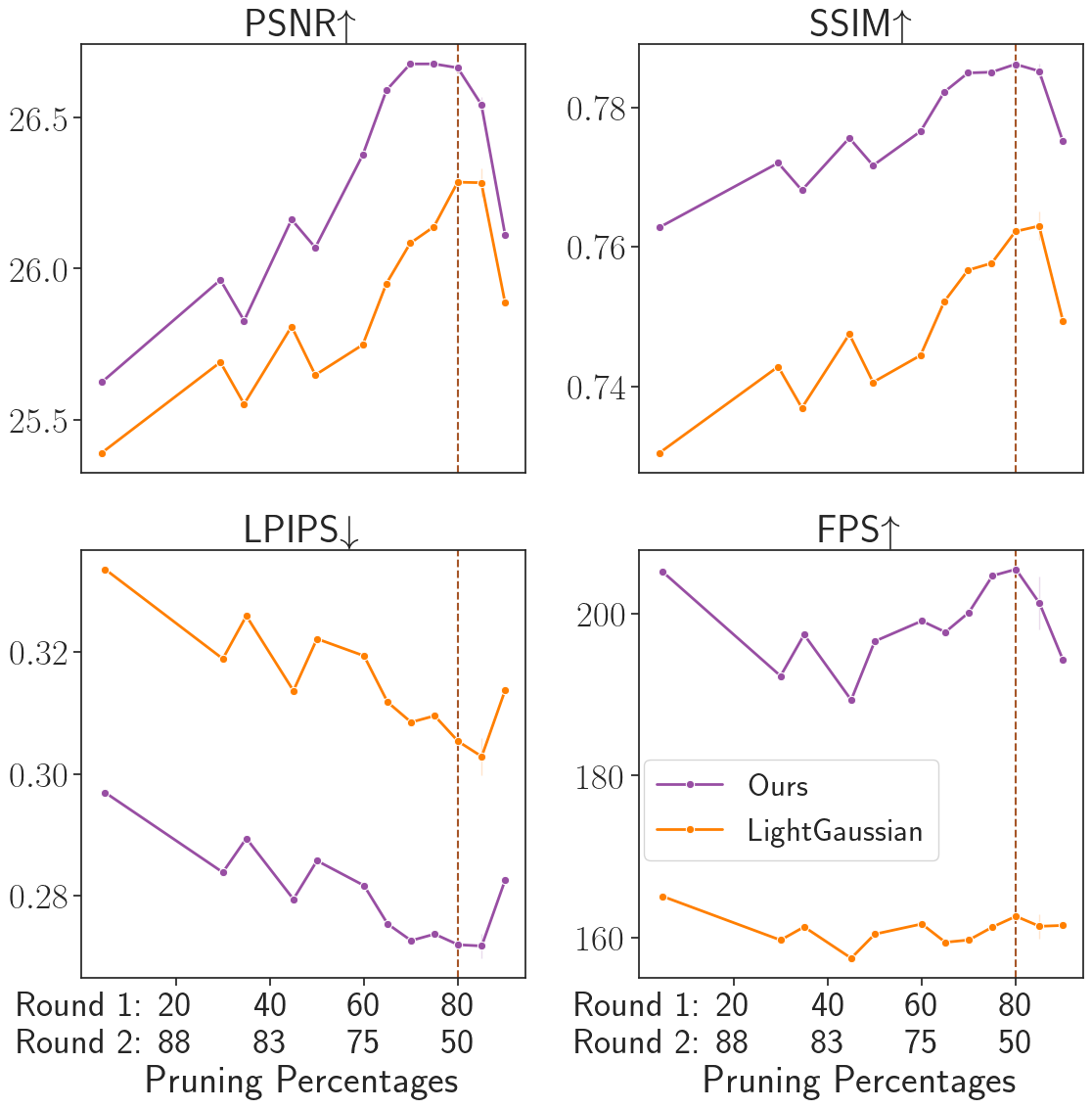}
  \vspace{-5mm}
  \caption{The average PSNR, SSIM, LPIPS, and FPS across the MipNeRF-360 dataset scene after pruning approximately $90\%$ of Gaussians with different per-round percentages in our two-round pipeline. The dotted red line denotes our chosen per-round pruning percentages of $80\%$ then $50\%$.
  }
  \vspace{-4mm}
  \label{fig:90_plot_mipnerf}
\end{figure}

Figure~\ref{fig:90_plot_mipnerf} plots the average metrics for each permutation of per-round pruning percentages that results in approximately $90\%$ total pruning across all scenes in the Mip-NeRF 360 dataset. Similar to the \textit{bicycle} scene evaluated in  Figure~\ref{fig:90_plot}, pruning $80\%$ of Gaussians in the first round and then $50\%$ in the second optimizes image quality and rendering speed at exactly $90\%$ total pruning. Our method outperforms LightGaussian across all metrics and per-round pruning percentage permutations.

\subsection{Scene Evaluations}
\label{sec:appendix:scene_eval}

PSNR, SSIM, LPIPS, and FPS for each scene from the Mip-NeRF 360, Tanks\&Temples, and Deep Blending datasets that was used in 3D-GS~\cite{kerbl3Dgaussians} are recorded in Tables~\ref{tab:psnr_scenes}, \ref{tab:ssim_scenes}, \ref{tab:lpips_scenes}, and~\ref{tab:fps_scenes}, respectively. 
Note that the sizes of the pruned scenes in this section are identical because exactly $90\%$ of Gaussians were removed from each of them using our two step prune-refine method. FPS is collected using a Nvidia RTXA4000 GPU.

\begin{table*}[t!]
\caption{PSNR on each scene after two steps of prune-refine.}
\centering
\resizebox{\linewidth}{!}{
\begin{tabular}{l*{9}{c}|*{2}{c}|*{2}{c}}
    \toprule
    \multirow{2}{*}{\textbf{Methods}} & \multicolumn{9}{c}{Mip-NeRF 360} & \multicolumn{2}{c}{Tanks\&Temples} & \multicolumn{2}{c}{Deep Blending} \\
    \cmidrule{2-14}
     & bicycle & bonsai & counter & flowers & garden & kitchen & room & stump & treehill & train & truck & drjohnson & playroom \\
    \midrule
  Baseline (3D-GS) & 25.09 & 32.25 & 29.11 & 21.34 & 27.28 & 31.57 & 31.51 & 26.55 & 22.56 & 22.10 & 25.43 & 28.15 & 29.81 \\
 LightGaussian & 24.34 & 29.64 & 27.57 & 20.70 & 25.78 & 29.45 & 30.65 & 25.88 & \textbf{22.49} & \textbf{21.35} & \textbf{24.81} & 27.73 & 29.29 \\
 Ours & \textbf{24.72} & \textbf{30.64} & \textbf{28.00} & \textbf{20.86} & \textbf{26.23} & \textbf{29.83} & \textbf{31.03} & \textbf{26.30} & 22.39 & 21.03 & 24.40 & \textbf{28.00} & \textbf{29.71} \\
    \midrule
    \bottomrule
\end{tabular}
}
\label{tab:psnr_scenes}
\end{table*}

\begin{table*}[h!]
\caption{SSIM on each scene after two steps of prune-refine.
}
\centering
\resizebox{\linewidth}{!}{
\begin{tabular}{l*{9}{c}|*{2}{c}|*{2}{c}}
    \toprule
    \multirow{2}{*}{\textbf{Methods}} & \multicolumn{9}{c}{Mip-NeRF 360} & \multicolumn{2}{c}{Tanks\&Temples} & \multicolumn{2}{c}{Deep Blending} \\
    \cmidrule{2-14}
     & bicycle & bonsai & counter & flowers & garden & kitchen & room & stump & treehill & train & truck & drjohnson & playroom \\
    \midrule
    Baseline (3D-GS) & 0.7467 & 0.9457 & 0.9140 & 0.5875 & 0.8558 & 0.9317 & 0.9255 & 0.7687 & 0.6352 & 0.8134 & 0.8782 & 0.8778 & 0.8854 \\
    LightGaussian & 0.6801 & 0.8921 & 0.8562 & 0.5343 & 0.7833 & 0.8830 & 0.8989 & 0.7256 & 0.5956 & 0.7349 & 0.8512 & 0.8546 & 0.8747 \\
    Ours & \textbf{0.7270} & \textbf{0.9261} & \textbf{0.8917} & \textbf{0.5548} & \textbf{0.8189} & \textbf{0.9128} & \textbf{0.9152} & \textbf{0.7570} & \textbf{0.6248} & \textbf{0.7600} & \textbf{0.8541} & \textbf{0.8762} & \textbf{0.8861}\\
    \midrule
    \bottomrule
\end{tabular}
}
\label{tab:ssim_scenes}
\end{table*}

\begin{table*}[h!]
\caption{LPIPS on each scene after two steps of prune-refine.
}
\centering
\resizebox{\linewidth}{!}{
\begin{tabular}{l*{9}{c}|*{2}{c}|*{2}{c}}
    \toprule
    \multirow{2}{*}{\textbf{Methods}} & \multicolumn{9}{c}{Mip-NeRF 360} & \multicolumn{2}{c}{Tanks\&Temples} & \multicolumn{2}{c}{Deep Blending} \\
    \cmidrule{2-14}
     & bicycle & bonsai & counter & flowers & garden & kitchen & room & stump & treehill & train & truck & drjohnson & playroom \\
    \midrule
     Baseline (3D-GS) & 0.2442 & 0.1811 & 0.1838 & 0.3602 & 0.1225 & 0.1165 & 0.1973 & 0.2429 & 0.3460 & 0.2077 & 0.1476 & 0.2895 & 0.2823 \\
     LightGaussian & 0.3338 & 0.2568 & 0.2801 & 0.4258 & 0.2407 & 0.2042 & 0.2625 & 0.3132 & 0.4315 & 0.3227 & 0.2041 & 0.3383 & 0.3201 \\
     Ours & \textbf{0.2965} & \textbf{0.2281} & \textbf{0.2297} & \textbf{0.4211} & \textbf{0.1997} & \textbf{0.1545} & \textbf{0.2278} & \textbf{0.2836} & \textbf{0.4062} & \textbf{0.2967} & \textbf{0.1916} & \textbf{0.3067} & \textbf{0.2963} \\
    \midrule
    \bottomrule
\end{tabular}
}
\label{tab:lpips_scenes}
\end{table*}

\begin{table*}[h!]
\caption{FPS on each scene after two steps of prune-refine. Results were collected with a Nvidia RTXA4000 GPU.
}
\centering
\resizebox{\linewidth}{!}{
\begin{tabular}{l*{9}{c}|*{2}{c}|*{2}{c}}
    \toprule
    \multirow{2}{*}{\textbf{Methods}} & \multicolumn{9}{c}{Mip-NeRF 360} & \multicolumn{2}{c}{Tanks\&Temples} & \multicolumn{2}{c}{Deep Blending} \\
    \cmidrule{2-14}
     & bicycle & bonsai & counter & flowers & garden & kitchen & room & stump & treehill & train & truck & drjohnson & playroom \\
    \midrule
     Baseline (3D-GS) & 32.32 & 106.65 & 79.70 & 66.48 & 37.60 & 64.46 & 76.31 & 54.04 & 59.08 & 114.45 & 81.26 & 56.79 & 76.79 \\
     LightGaussian & 110.94 & 208.15 & 168.48 & 182.28 & 133.77 & 165.98 & 172.46 & 169.98 & 147.03 & 378.50 & 279.55 & 206.94 & 261.27 \\
     Ours & \textbf{127.85} & \textbf{289.21} & \textbf{222.77} & \textbf{205.21} & \textbf{164.79} & \textbf{237.66} & \textbf{231.63} & \textbf{179.38} & \textbf{184.78} & \textbf{494.76} & \textbf{287.44} & \textbf{284.81} & \textbf{318.06} \\
    \midrule
    \bottomrule
\end{tabular}
}
\label{tab:fps_scenes}

\end{table*}

\newpage
\newpage

\subsection{Prune-Refining EAGLES}

\label{sec:appendix:eagles}
In Table~\ref{tab:eagles}, we ablate our \acronym{} pipeline against LightGaussian's pipeline on an EAGLES~\cite{girish2023eagles} model of the Mip-NeRF 360 \textit{bicycle} scene. Notice that the base EAGLES model produces similar metrics to vanilla 3D-GS despite being $2.51\times$ smaller than it. By applying \acronym{} to the EAGLES model, we further reduce its size to $25.14\times$ smaller than the vanilla 3D-GS model while achieving better image quality and rendering speed than LightGaussian.

\begin{table}[h]
\caption{Results from training EAGLES~\cite{girish2023eagles} on the \emph{bicycle} scene and then running two steps of prune-refine with our and LightGaussian's methods. }
\centering
\resizebox{\columnwidth}{!}{
\begin{tabular}{l*{5}{c}}
    \toprule
    \multirow{2}{*}{\textbf{Methods}} & \multicolumn{5}{c}{Mip-NeRF 360 Bicycle Scene} \\
    \cmidrule{2-6} & PSNR$\uparrow$ & SSIM$\uparrow$ & LPIPS$\downarrow$ & FPS$\uparrow$ & Size (MB)$\downarrow$ \\
    \midrule
    3D-GS        & 25.09 & 0.7467 & 0.2442 & 32.12 & 1345.58 \\
    Baseline (EAGLES) & 25.07 & 0.7508 & 0.2433 & 47.82 & 535.21 \\
    EAGLES + LightGaussian  & 23.57 & 0.6082 & 0.4039 & 109.26 & 53.52 \\
    EAGLES + Ours & \textbf{24.01} & \textbf{0.6686} & \textbf{0.3566} & \textbf{144.56} & 53.52
 \\
    \midrule
    \bottomrule
\end{tabular}}
\label{tab:eagles}
\end{table}

\subsection{Comparison with Other Pruning Methods}
\label{sec:appendix:other_pruning}
Several other papers also introduce pruning techniques. Compact-3DGS\cite{lee2023compact} reports that they prune $58.7\%$ of Gaussians from the Mip-NeRF 360 \textit{bonsai} scene, slightly improving PSNR from $29.87$ to $29.91$. After pruning $58.7\%$ of Gaussians with a single round of prune-refine, our method produces a much larger PSNR boost from $32.25$ to $32.55$. Mini-Splatting~\cite{fang2024minisplattingrepresentingscenesconstrained} and RadSplat~\cite{niemeyer2024radsplatradiancefieldinformedgaussian} achieve higher rendering quality through orthogonal methods like improved densification and pretrained NeRF models. Since our pruning approach can be combined with these methods, it is unclear if a direct comparison with their results is useful. As reported in Table~\ref{tab:compare_methods}, our $10\times$ reduction in the number of Gaussians is significantly higher than the $2.28\times$ and $6.94\times$ reductions reported by Compact-3DGS and Mini-Splatting.

\subsection{Potential Directions for Future Research}
\label{sec:appendix:next_steps}

The theory behind our Hessian approximation relies on a set of views whose L1 loss is close to zero, so we compute our spatial sensitivity score on the training views after optimization to remain as mathematically principled as possible. However, our score (1) does not rely on ground-truth data and (2) can be computed when the L1 loss is sufficiently low earlier in the 3D-GS training pipeline. These properties may provide a useful first step for several potential directions for future research.

Further optimizations and/or sufficiently powerful hardware could allow our sensitivity score to be used to prune the model during training. This is a promising research direction that could potentially lead to even higher compression and rendering speeds. Additionally, it may be possible to identify a subset of pixels, rays, or views that would produce a score that is most effective for pruning. Since this is a non-trivial, combinatorially difficult problem and our pruning score can be computed across the entire training set in seconds, we also leave this as an open research question.

Our score can be extended to other 3D Gaussian pipelines such as the Mini-Splatting and PGSR pipelines~\cite{fang2024minisplattingrepresentingscenesconstrained,chen2024pgsr}. Some considerations must be made, such as the effect of each Gaussian on the depth and normal maps produced by these approaches along with the rendered image. Nevertheless, the math in our main paper holds if we include the depth and normal maps as additional channels. Pruning anchor-based approaches like Scaffold-GS~\cite{lu2023scaffold} and derived works like Octree-GS~\cite{ren2024octree} is a more difficult problem. Directly pruning Gaussians with our current pipeline will not work because a fixed number of Gaussians are produced for each anchor point; pruning anchor points is ill-advised because they are generated in areas of the scene that do not have sufficient geometry. While our PUP 3D-GS approach presents a method for directly quantifying a Gaussian's importance in the scene, determining how to map that score to pruning anchors is an open research problem.

FisherRF~\cite{jiang2023fisherrf} also computes Fisher information for 3D Gaussian Splats, but uses it to perform active-view selection and post-hoc uncertainty visualization instead of pruning. Furthermore, FisherRF only approximates the \emph{diagonal} of the Fisher matrix and uses the color parameters of the Gaussians, whereas our approach uses the spatial mean and scaling parameters to compute a more accurate \emph{block-wise} approximation. Our sensitivity pruning score can be directly repurposed for these applications, but we leave this to future work because it is not the focus of our paper.

\subsection{Additional Scene Visualizations}
\label{sec:appendix:scene_visuals}
\begin{figure*}[t!]
  \includegraphics[width=\linewidth]{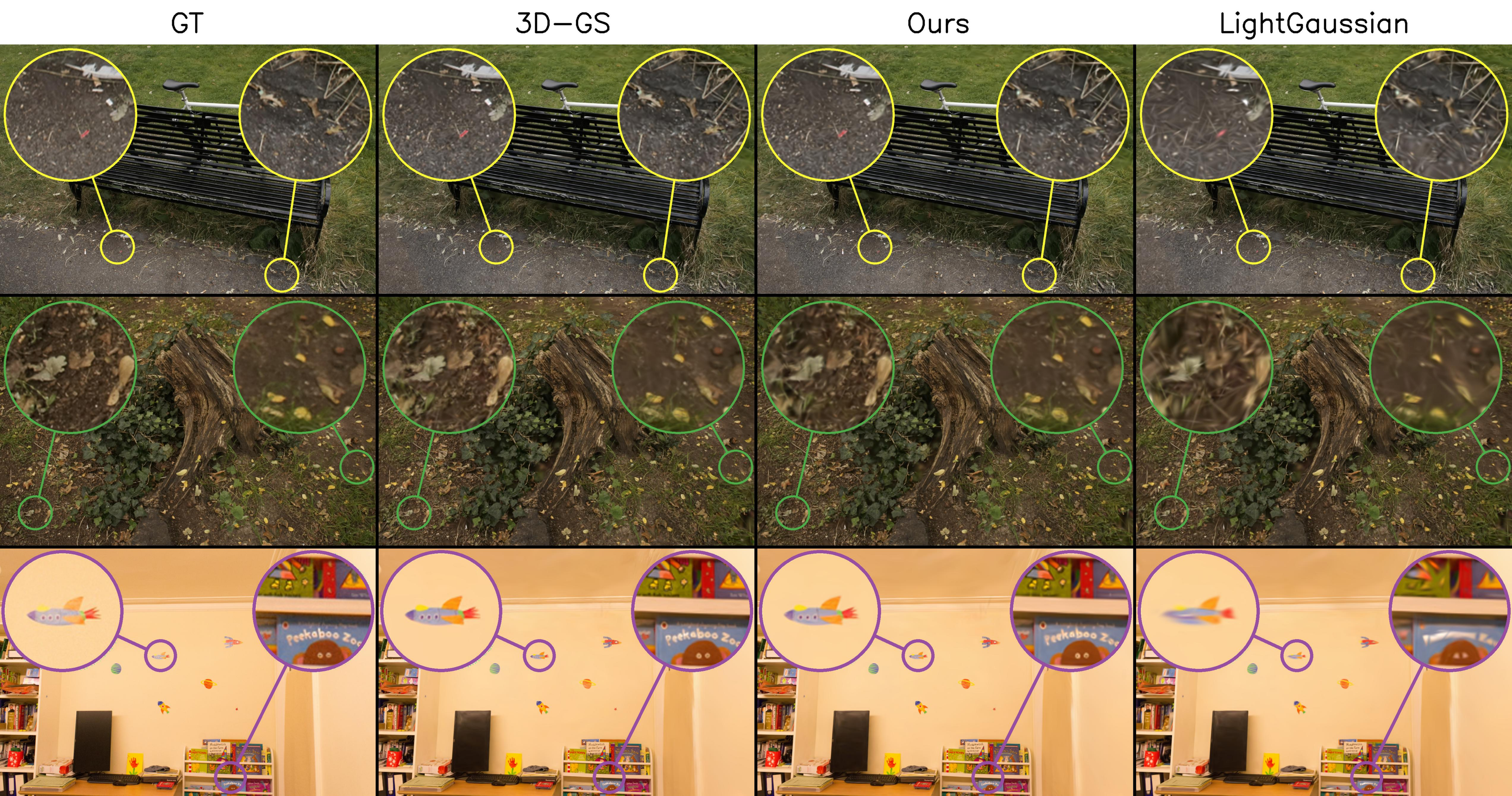}
  \caption{Visual comparison after two rounds of prune-refine using our method and LightGaussian's method. Top: \emph{bicycle} from the Mip-Nerf 360 dataset. Middle: \emph{stump} from the Mip-Nerf 360 dataset. Bottom: \emph{playroom} from the Deep Blending dataset. A larger example image of \emph{playroom} can be found in Figure~\ref{fig:teaser}.}
  \label{fig:appendix_figure}
\end{figure*}

Figure~\ref{fig:appendix_figure} provides a visual comparison of the ground truth image against renderings from the 3D-GS model before and after pruning with our \acronym{} and LightGaussian's pipelines. Notice that our method consistently achieves higher visual fidelity and retains more salient foreground information like individual leaves and legible text.

\subsection{Background Degradation}
\label{sec:appendix:background_degradation}

At the extreme pruning ratios used in our work, we observe some background degradation in pruned models. Intuitively, background regions that are observed from fewer viewpoints exhibit higher uncertainty than well-observed foreground regions, making them more susceptible to pruning. Given our aggressive pruning threshold of 90\%, lossless compression is not expected. However, our method prioritizes preserving fine details in the foreground while pruning less important Gaussians in the background to maintain overall visual quality. We do not consider this a failure case because retaining foreground details over background details is often preferable.

\begin{figure*}[b!]
  \includegraphics[width=\linewidth]{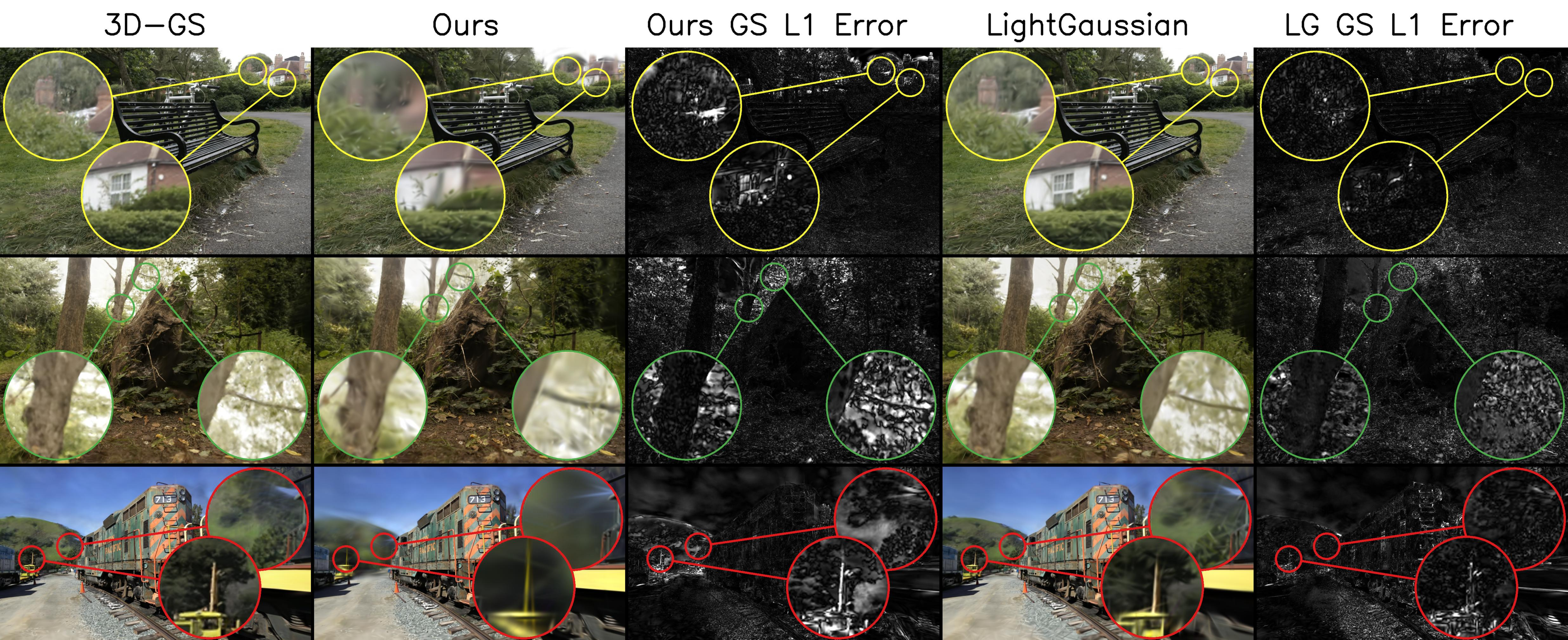}
  \caption{Cases where our method produces more background L1 error against renderings from the base 3D-GS model than LightGaussian.
  Columns "3D-GS", "Ours", and "LightGaussian" are the same as in Figure~\ref{fig:experiments_figure}.
  Columns "Ours GS L1 Error" and "LG GS L1 Error" are the L1 error images of our method and LightGaussian against the original 3D-GS reconstruction of the scene.
  Our method prioritizes retaining visual quality and fine details in the foreground over preserving less important information in the background.
  }
  \label{fig:failure_cases}
\end{figure*}

In contrast, LightGaussian exhibits the opposite tendency, often favoring background preservation over foreground fidelity. Figure~\ref{fig:failure_cases} illustrates this difference by comparing L1 residuals against renderings from the base 3D-GS model and highlighting instances where PUP 3D-GS degrades the background more than LightGaussian. A potential strategy for mitigating this trade-off is to reweight the background Gaussian scores using masking, incorporating a tunable parameter to control the balance between foreground and background degradation. We leave this exploration to future work.

\subsection{Scene Residuals}
\label{sec:appendix:l1_gt_error}
Figures~\ref{fig:l1_gt_error} and \ref{fig:l1_gs_error} provide visual comparisons of the L1 residual of renderings from the 3D-GS model before and after pruning $90\%$ of Gaussians with our \acronym{} and LightGaussian's pipelines. Figure~\ref{fig:l1_gt_error} compares the L1 residuals with respect to the ground truth image, while Figure~\ref{fig:l1_gs_error} compares them with respect to renderings from the base 3D-GS model. In both cases, our \acronym{} pipeline produces less L1 error and retains more salient foreground information than LightGaussian’s. 

\clearpage
\clearpage

\begin{figure*}[t]
  \includegraphics[width=\linewidth]{figures/l1_gt_error_figure.pdf}
  \caption{
  Visual comparison after two rounds of prune-refine using our method and LightGaussian's with additional L1 error visualizations against the ground truth images.
  Columns "GT", "Ours", and "LightGaussian" are the same as in Figure~\ref{fig:experiments_figure}.
  Columns "Ours L1 Error" and "LG L1 Error" are the L1 Error images of our method and LightGaussian against the ground truth images. Our method produces lower error than LightGaussian in all examples.
  }
  \label{fig:l1_gt_error}
\end{figure*}

\begin{figure*}[t]
  \includegraphics[width=\linewidth]{figures/l1_gs_error_figure.pdf}
  \caption{
  Visual comparison after two rounds of prune-refine using our method and LightGaussian's with additional L1 error visualizations against renderings from the base 3D-GS model.
  Columns "3D-GS", "Ours", and "LightGaussian" are the same as in Figure~\ref{fig:experiments_figure}.
  Columns "Ours GS L1 Error" and "LG GS L1 Error" are the L1 error images of our method and LightGaussian against the original 3D-GS reconstruction of the scene. Our method produces lower error than LightGaussian in all examples.}
  \label{fig:l1_gs_error}
\end{figure*}

\end{document}